\pdfoutput=1

\documentclass[11pt]{article}

\usepackage[final]{acl}

\usepackage{times}
\usepackage{latexsym}

\usepackage[T1]{fontenc}

\usepackage[utf8]{inputenc}

\usepackage{microtype}
\usepackage{booktabs}
\usepackage{multirow}
\usepackage{inconsolata}
\usepackage[dvipsnames]{xcolor}
\usepackage{graphicx}
\usepackage{amsmath} 
\usepackage{subfigure}
\usepackage{algorithm}
\usepackage{algpseudocode}
\usepackage{fancyhdr}

\definecolor{darkred}{RGB}{200,50,50}
\definecolor{darkgreen}{RGB}{30,150,30}
\definecolor{darkblue}{RGB}{50,100,200}
\title{Mitigating Visual Knowledge Forgetting in MLLM Instruction-tuning via Modality-decoupled Gradient Descent}

\author{
Junda Wu$^1$, Yuxin Xiong$^1$, Xintong Li$^1$, Yu Xia$^1$, Ruoyu Wang$^2$, Yu Wang$^1$ \\
\textbf{Tong Yu$^3$, Sungchul Kim$^3$, Ryan Rossi$^3$, Lina Yao$^2$, Jingbo Shang$^1$, Julian McAuley$^1$}
 \\
$^1$UC San Diego \quad 
$^2$University of New South Wales \quad 
$^3$Adobe Research \\
\texttt{\{juw069,y7xiong,xil240,yux078,yuw164,jshang,jmcauley\}@ucsd.edu} \\
\texttt{\{ruoyu.wang5,lina.yao\}@unsw.edu.au} \quad \texttt{\{tyu,sukim,ryrossi\}@adobe.com} \\
}

\begin{document}



\maketitle

\begin{abstract}
Recent MLLMs have demonstrated strong visual understanding and reasoning after large-scale multimodal pre-training. However, instruction-tuning is typically text-driven with limited visual supervision, leading to significant visual forgetting and degradation of pre-trained visual knowledge. Existing fine-tuning and continual learning methods compress visual representations and emphasize task alignment over visual retention, failing to address this challenge.
We present a novel perspective using effective rank to quantify the loss of visual representation richness, framing visual forgetting as excessive compression under the information bottleneck principle. To address this, we propose modality-decoupled gradient descent (MDGD), which regulates gradient updates to preserve the effective rank of visual features and explicitly disentangles visual learning from task-specific alignment. We further introduce a memory-efficient fine-tuning variant using gradient masking for parameter-efficient adaptation.
Extensive experiments show that MDGD effectively mitigates visual forgetting across downstream tasks and models, maintaining pre-trained visual knowledge while supporting strong task adaptation.
\end{abstract}

%
\section{Introduction}
\begin{figure}[btp]
    \centering
    \subfigure[LLaVA on OKVQA]{%
        \includegraphics[width=0.47\linewidth]{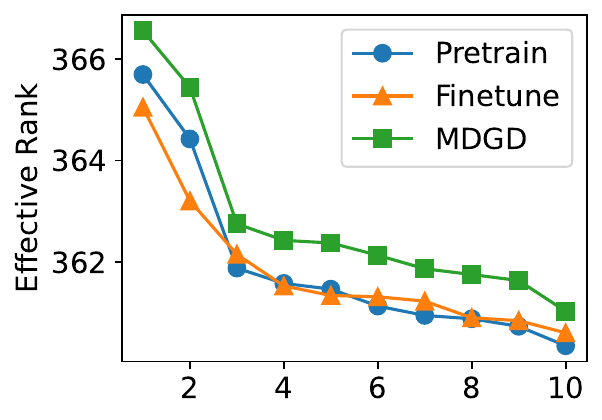} 
        \label{fig:erank-token1}
    }
    \hfill
    \subfigure[LLaVA on POPE]{%
        \includegraphics[width=0.47\linewidth]{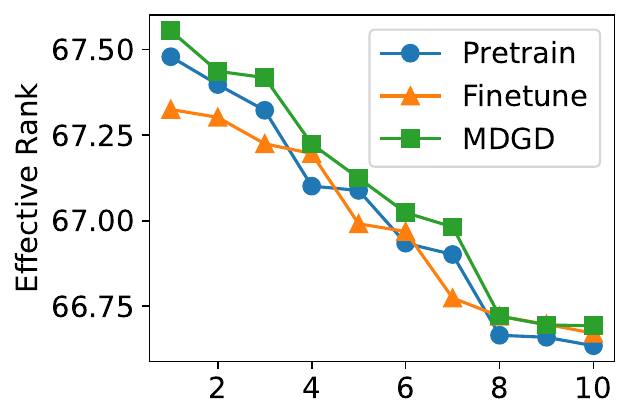} 
        \label{fig:erank-token2}
    }
    
    
    \subfigure[MiniCPM on PathVQA]{%
        \includegraphics[width=0.48\linewidth]{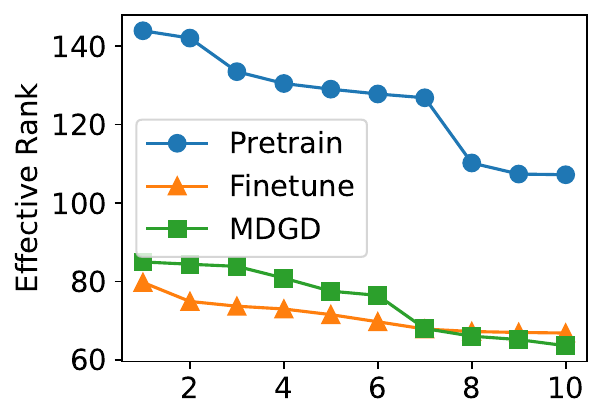} 
        \label{fig:erank-token3}
    }
    \hfill
    \subfigure[MiniCPM on POPE]{%
        \includegraphics[width=0.48\linewidth]{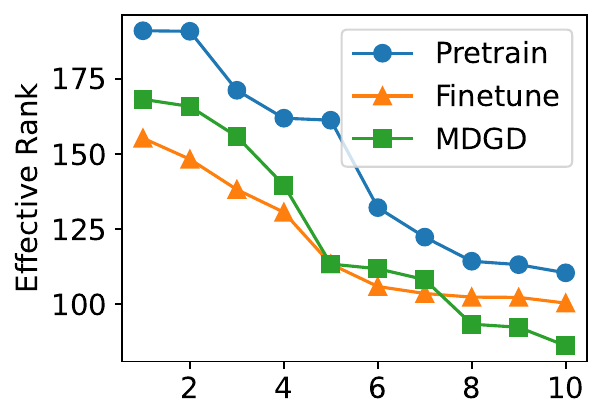} 
        \label{fig:erank-token4}
    }

    
    \caption{
    The top-10 image tokens with the highest effective ranks on OKVQA and POPE encoded by LLaVA, and PathVQA and POPE encoded by MiniCPM. 
    We compare pretrained, finetuned, and MDGD-finetuned models. 
    Effective rank \cite{wei2024diff} quantifies representation richness, which show that MDGD preserves higher effective rank, mitigating visual forgetting.
    }
    \label{fig:intro}
    \vspace{-1em}
\end{figure}
Multimodal large language models (MLLMs) enhanced visual understanding and reasoning by pre-training on large-scale multimodal datasets 
with comprehensive visual descriptions that integrate textual and visual knowledge~\cite{liu2024visual,yao2024minicpmvgpt4vlevelmllm,li2023blip,bai2023qwen,liu2024multimodal,wu2024visual}. 
These models achieve strong performance across various vision-language tasks, 
such as visual question answering \cite{jin2024rjua,wu2025doc}, multimodal reasoning \cite{zhang2024large,jiang2024killing,yan2024list}, multimodal recognition \cite{shenoy2024lumos,wu2024visual}, 
and personalized multimodality \cite{wu2024personalized,wu2025pdb,huang2025towards}.
However, adapting pre-trained MLLMs to downstream tasks via instruction-tuning \cite{wu2024commit,li2024vision,li2023fine,panagopoulou2023x,liu2024multimodal} presents a critical challenge of visual forgetting. 
Unlike pre-training, where models receive rich visual-text alignment, instruction-tuning is often text-driven with limited direct visual supervision. 
This shift in training focus leads to the degradation of pre-trained visual encoding \cite{zhou2024mitigating,niu2024text,wu2024commit,ko2023large}, 
negatively impacting model generalizability across downstream tasks that require strong visual knowledge \cite{bai2024hallucination,huang2024visual}. 
Addressing this challenge is essential for ensuring MLLMs retain their visual capabilities while aligning with new tasks efficiently.

While several approaches have attempted to mitigate catastrophic forgetting in neural networks through direct fine-tuning and continual learning methods \cite{shi2024continual, wu2024continual, zhu2024model, zheng2024beyond}, 
these methods often overlook the unique challenge of preserving visual knowledge in multimodal large language models (MLLMs). 
Directly fine-tuning MLLMs on new tasks often leads to overfitting to textual instructions while inadvertently suppressing visual representations \citep{zhai2023investigating}.
Existing continual learning strategies, such as regularization and replay methods, tend to focus on retaining language-based knowledge, 
neglecting the trade-off between compressing visual representations and aligning them with task-specific instructions \cite{zhou2024mitigating, niu2024text, wu2024commit, ko2023large}, 
leading to the degradation of pre-trained visual knowledge.
Task-orthogonal gradient descent techniques have shown promise in disentangling gradients for multi-task optimization.
However, their practical application in MLLMs poses unique challenges. 
MLLMs are pre-trained on vast and heterogeneous multimodal datasets \cite{liu2024visual, li2023blip, bai2023qwen}, 
where it is challenging to isolate task-specific gradients, 
causing the components critical for visual understanding to become entangled with other features.

To gain a fundamental view of the challenge of visual knowledge forgetting in MLLM instruction tuning, 
we adopt an information bottleneck (IB) perspective that characterizes the trade-off between retaining input information and ensuring output predictiveness \cite{tishby2000information}.
To investigate the degradation of crucial pre-trained visual knowledge, we introduce a novel perspective that leverages effective rank to quantify the richness of the encoded visual representation from MLLMs.
Specifically, we illustrate the visual forgetting problem in Figure~\ref{fig:intro}, where we observe a consistent effective rank reduction problem caused by MLLM instruction tuning.
Based on this view, we propose a modality-decoupled gradient descent (MDGD) method, which disentangles the optimization of visual understanding from task-specific alignment,
MDGD regulates gradient updates to maintain the effective rank of visual representations compared with pre-trained MLLMs,
while mitigating the over-compression effects described by the information bottleneck. 
Intuitively, visual forgetting occurs due to the shift from rich multimodal pre-training to instruction-tuning, 
where text-based supervision dominates without direct visual supervision. 
By explicitly decoupling the task-specific alignment with visual representation learning, MDGD preserves expressive and robust visual features. 
To further improve efficiency in instruction-tuning, we introduce a memory-efficient fine-tuning strategy using gradient masking, 
which selectively updates a subset of model parameters for parameter-efficient fine-tuning (PEFT).
This approach reduces computational overhead while ensuring that crucial pre-trained visual representations are retained.

We summarize our contributions as follows: 
\begin{itemize}
\item We analyze the visual knowledge forgetting problem in MLLM instruction tuning and frame the problem through the lens of effective rank and information bottleneck theory.
\item We propose MDGD, which decouples visual optimization from task-specific alignment to preserve visual representations and introduces a PEFT variant MDGD-GM to reduce computational overhead through gradient masking. 
\item We conduct comprehensive experiments on various MLLMs and downstream tasks, demonstrating that MDGD effectively mitigates visual forgetting while enabling strong adaptation to new tasks.
\end{itemize}

\subsection{Visual Knowledge Forgetting in MLLMs}
Catastrophic forgetting, where a model loses previous knowledge while learning new tasks, is a major challenge in continual learning~\cite{wang2023orthogonal}. This problem is now widely recognized in LLMs and MLLMs~\cite{wu2024continual,luo2023empirical,zhai2023investigating}. Although various methods, including fine-tuning, task-orthogonal gradient descent, knowledge distillation, and replay, have been adapted to mitigate forgetting~\cite{shi2024continual,wu2024continual,zhu2024model,zheng2024beyond}, they often fail to preserve rich visual features. Fine-tuning on new tasks tends to overfit text and suppress visual information, while parameter-efficient methods like LoRA also suffer from forgetting~\cite{fawi2024curlora,liu2024learning}. Model Tailor~\cite{zhu2024model} adapts the LLM backbone but does not address visual knowledge forgetting, which can lead to hallucination or degraded generalization~\cite{zhai2023investigating}. In contrast, our approach synchronizes the training of the visual encoder and LLM, preserving pre-trained visual knowledge during instruction tuning.

\subsection{Information Theory in LLMs}
The Information Bottleneck (IB) principle~\cite{tishby2000information} has been used in LLMs to compress input while retaining task-relevant information~\cite{deletang2023language,valmeekam2023llmzip,wei2024diff,wu2022context}. Prior works use IB to extract robust features~\cite{zhang2022improving,wu2024infoprompt} and enable feature attribution~\cite{li2022explanation,jiang2020inserting}, but these focus on language models, not multimodal settings~\cite{yang2025exploring}. Existing information-theoretic transfer learning methods~\cite{tseng2024semantic,wu2024infoprompt,ling2024convergence} also do not address the unique challenges of MLLMs, where modalities are deeply entangled. Our method instead uses effective rank to measure and counteract visual representation compression. The proposed MDGD explicitly decouples visual learning from task alignment, going beyond previous IB-based approaches.

\section{Preliminary}\label{sec:prelim}
\noindent\textbf{Task Definition.}
Given an MLLM $\pi_{\theta}$ and instruction-tuning dataset $D$, the image prompt $I\in \Omega$ is encoded by a visual encoder $f$ 
into a sequence of $M$ visual tokens $f(I)=X^v=(x_1^v,x_2^v,\dots,x_M^v)$.
During instruction tuning,
the textual instructions $T\in D$ are tokenized as $X^l=(x_1^l,x_2^l,\dots,x_N^l)$ using the tokenizer of the backbone LLM, 
which is querying the MLLM to generate textual responses conditioned on the multimodal inputs,
\begin{equation}
    \hat{y}_k \sim \pi_{\theta}(\cdot \mid X^v, X^l, y_{<k}).
\end{equation}
Therefore, the learning objective of visual instruction-tuning for $K$ samples is to 
maximize the average log-likelihood of the ground truth answer tokens $y =(y_1, y_2, \dots, y_T)$ of each sample,
\begin{equation}\label{eq:task_loss}
    \mathcal{L}_{vl}(\theta) = 
    -  \sum_{t=1}^{T} 
    \log \pi_\theta \left( y_t \mid X^v, X^l, y_{<t} \right),
\end{equation}
where multimodal instructions $X^v$ and $X^l$ both serve as generation conditions.

\vspace{1em}\noindent\textbf{An Information Bottleneck Perspective on Visual Knowledge Forgetting.}
In multimodal models, the information bottleneck~\cite{mai2022multimodal} (IB) framework provides a powerful lens to understand how representations are formed. 
In our setting, the IB principle seeks a representation \( Z \) that is maximally informative about the output \( y \) while discarding irrelevant details from the inputs. 
For an MLLM that processes visual inputs \( X^v \) and textual inputs \( X^l \), a full IB objective might take the form:
\begin{equation}\label{eq:ib_vision}
    \min_{\theta} \quad \mathcal{L}_{\text{IB}}^{\text{vision}}(\theta) = - I(y; Z) + \beta\, I(X^v; Z).
\end{equation}
where \( I(\cdot;\cdot) \) denotes mutual information and \(\beta\) controls the trade-off between predictive power and compression.
This formulation explicitly highlights the risk of discarding visual details when the model is optimized primarily to predict \( y \).
Based on the information bottleneck view, we further analyze the visual forgetting problem in Appendix~\ref{sec:visual_forget}.

\vspace{1em}\noindent\textbf{Effective Rank as a Measure of Representation Richness.}
To quantify the information content retained in a representation, we use the effective rank metric~\cite{roy2007effective}. Given a representation matrix \( Z \) whose singular values are \( \{\sigma_i\} \), the effective rank is defined as:
\begin{equation}\label{eq:erank}
    \text{erank}(Z) = \exp\Biggl(-\sum_{i} p_i \log p_i\Biggr),
\end{equation}
where $p_i = \sigma_i/\sum_j \sigma_j$.
This measure, based on the entropy of the singular value distribution, captures the “richness” or intrinsic dimensionality of \( Z \). A higher effective rank indicates that the representation spans a larger subspace, whereas a lower effective rank implies that the representation has been overly compressed.

\section{MDGD: Modality-Decoupled Gradient Regularization and Descent}
Motivated by the visual forgetting problem caused by the degradation of multimodal encoding in Eq.~\eqref{eq:erank_degradation}, we introduce a modality-decoupling gradient regularization (\textbf{MDGD}) to approximate orthogonal gradients between visual understanding drift and downstream task optimization. Specifically, leveraging modality-decoupled gradients $\Bar{g}_\theta$ and $\Bar{g}_\phi$ derived from the current MLLM and a pre-trained MLLM respectively, we propose a gradient regularization term $\Tilde{g}_\theta$ for more efficient multimodal instruction tuning, which promotes the alignment of downstream tasks while mitigating visual forgetting \cite{zhu2024model}. Since MDGD requires the estimation of parameter gradients, we could not directly apply parameter-efficient fine-tuning methods (\emph{e.g.}, LoRA \cite{hu2021lora}). Thus, we alternatively formulate the regularization as a gradient mask $M_{\Tilde{g}_\theta}$, which allows efficient fine-tuning only on a subset of masked model parameters.

\subsection{Modality Decoupling}
Based on the information bottleneck objective in Eq.~\eqref{eq:ib_vision}, the objective encourages the model to maximize $I(y; Z)$ while compressing $I(X^v; Z)$ \cite{tishby2000information, alemi2016deep}. 
In practice, this compression may discard useful visual details, leading to visual forgetting. To mitigate such compression and preserve the pre-trained visual knowledge, we follow the KL divergence loss
$D_{\text{KL}}\Bigl(\mu_\phi(X^v) \,\Big\|\, \pi_\theta (X^v)\Bigr)$
to constrain the current model’s visual representation $\pi_\theta(X^v)$ to remain close to the pre-trained distribution $\mu_\phi(X^v)$, 
thereby preserving the mutual information $I(X^v; Z)$ that would otherwise be reduced by the compression \cite{hinton2015distilling, lopez2018information}. 
However, since MLLMs cannot directly track the distributions of image tokens, we instead introduce an auxiliary loss function
\begin{equation}\label{eq:visual_loss}
    \mathcal{L}_v(\phi,\theta) = \|\mu(X^v|\phi) - \pi(X^v|\theta)\|_1,
\end{equation}
which approximates the KL divergence loss \cite{zhu2022wdibs,zhu2017unpaired} by penalizing discrepancies between the pre-trained visual representation and that obtained during instruction tuning. 

In the MLLM instruction tuning, the visual output tokens (e.g., $\{z^{vl}_k\}_{k=1}^M$) are encoded as latent representations. 
Such visual encoding cannot be directly supervised by any learning objective but is learned through textual gradient propagation of the negative log-likelihood loss in downstream tasks. 
To approximate the visual optimization direction, we derive the gradients of $\mathcal{L}_v(\phi,\theta)$ for both the pre-trained MLLM $\pi_\phi$ and the current MLLM $\pi_\theta$:
\begin{align*}
    h_{\phi} &= \nabla_{\phi}\mathcal{L}_v(\phi) = \boldsymbol{\lambda}(\phi,\theta) \cdot \nabla_\phi \mu(X^v|\phi), \\
    h_{\theta} &= \nabla_{\theta}\mathcal{L}_v(\theta) = -\boldsymbol{\lambda}(\phi,\theta) \cdot \nabla_\theta \pi(X^v|\theta),
\end{align*}
where $\boldsymbol{\lambda}(\phi,\theta) = \text{sign}\left( \mu(X^v|\phi) - \pi(X^v|\theta) \right)$.
Intuitively, when the MLLM's visual understanding drift causes visual forgetting, we further derive the orthogonal task gradients $\Bar{g}_\phi$ and $\Bar{g}_\theta$:
\begin{align}\label{eq:orth}
    \Bar{g}_\phi &= \nabla_{\phi}\mathcal{L}_{vl}(\phi) - \frac{\nabla_{\phi}\mathcal{L}_{vl}(\phi)^\top h_{\phi}}{\|h_{\phi}\|^2} \cdot h_{\phi}, \\
    \Bar{g}_\theta &= \nabla_{\theta}\mathcal{L}_{vl}(\theta) - \frac{\nabla_{\theta}\mathcal{L}_{vl}(\theta)^\top h_{\theta}}{\|h_{\theta}\|^2} \cdot h_{\theta},
\end{align}
which enables \textbf{modality decoupling} of the downstream task loss gradient in Eq.\eqref{eq:task_loss} orthogonal to the visual understanding drift
for the pretrained MLLM $\Bar{g}_{\phi} \perp h_{\phi}$ and current MLLM $\Bar{g}_{\theta} \perp h_{\theta}$.

\begin{algorithm}[ht]
\caption{MDGD: Modality Decoupled Gradients Descent }
\label{alg}
\begin{algorithmic}[1]
\State \textbf{Inputs:} 
Pre-trained MLLM $\mu_\phi$, current MLLM $\pi_\theta$, instruction-tuning dataset $D$, and learning rate $\eta$
\State \textbf{Outputs:} The optimized model weights of $\pi_\theta$
\State \textbf{Initialize} $\pi_\theta \leftarrow \mu_\phi$
\For{Receive minibatch $D_i\subset D$}
    \State Calculate $\mathcal{L}_{vl}(\phi)$ of $\mu_\phi$, based on Eq.\eqref{eq:task_loss};
    \State Calculate $\mathcal{L}_{vl}(\theta)$ of $\pi_\theta$, based on Eq.\eqref{eq:task_loss};
    \State Extract visual encodings of $\mu(X^v|\phi)$;
    \State Extract visual encodings of $\pi(X^v|\theta)$;
    \State Calculate $\mathcal{L}_v(\phi,\theta)$, based on Eq.\eqref{eq:visual_loss};
    \State Derive orthogonal task gradients $\Bar{g}_\phi$ and $\Bar{g}_\theta$, according to Eq.\eqref{eq:orth};
    \If{Parameter-efficient fine-tuning}
        \State Calculate $M_{\Tilde{g}_\theta}$,based on Eq.\eqref{eq:masking};
        \State Update the model following Eq.\eqref{eq:opt-mask}.
    \Else
        \State Calculate $\Tilde{g}_\theta$, based on Eq.\eqref{eq:gd};
        \State Update the model following Eq.\eqref{eq:opt-gd}.
    \EndIf
\EndFor
\end{algorithmic}
\end{algorithm}

\subsection{Regularized Gradient Descent}
The auxiliary loss in Eq.~\eqref{eq:visual_loss} preserves the visual representation at a distribution level via the feature alignment auxiliary loss in Eq.~\eqref{eq:visual_loss}. 
However, the information bottleneck framework indicates that the gradient component compressing $I(X^v; Z)$ (\emph{i.e.}, $\nabla_\theta I(X^v; Z)$), 
can harm visual preservation by reducing the effective rank of the features \cite{achille2018information,lee2021compressive}.

To address this compression-induced drift, we incorporate an orthogonal gradient as a regularize. 
Motivated by multi-task orthogonal gradient optimization \cite{yu2020gradient, zhu2022gradient, dong2022gdod}, 
we leverage the gradient $\Bar{g}_\phi$ from the pre-trained model $\mu_\phi$, which reflects the accumulated visual drift and approximates a global orthogonal learning effect in the downstream task. 
We then project the current model’s gradient onto this direction:
\begin{equation}\label{eq:gd}
    \Tilde{g}_\theta = \frac{\Bar{g}_\theta^\top \Bar{g}_\phi}{\|\Bar{g}_\phi\|^2}\cdot \Bar{g}_\phi.
\end{equation}

In addition, to prevent discrepancies between the regularization and task gradients, we include the feature alignment auxiliary loss (Eq.~\eqref{eq:visual_loss}) in the overall objective. The final parameter update is:
\begin{equation}\label{eq:opt-gd}
    \pi_\theta \leftarrow \pi_\theta - \nabla_\theta\mathcal{L}_{vl}(\theta) - \nabla_\theta\mathcal{L}_v(\theta) - \Tilde{g}_\theta.
\end{equation}

\subsection{Enabling Parameter-efficient Fine-tuning of MDGD via Gradient Masking}
Parameter-efficient fine-tuning (PEFT) methods, such as adapters \cite{houlsby2019parameter} and LoRA \cite{hu2021lora}, aim to reduce the computational cost and memory usage when fine-tuning models on downstream tasks under practical constraints \cite{han2024parameter}. 
However, due to the requirement of directly estimating gradient directions on the pre-trained model parameters, MDGD cannot be directly applied to these PEFT methods, which introduce additional model parameters whose gradients are separate from the original model weights. 

To address this challenge, we propose a variant, MDGD-GM, by formulating the gradient regularization term in Eq.~\eqref{eq:gd} as gradient masking that selects model weights with efficient gradient directions. Specifically, we define it as
\begin{equation}\label{eq:masking}
    M_{\Tilde{g}_\theta} = \mathbf{1}\left\{\frac{\Bar{g}_\theta^\top \Bar{g}_\phi}{\|\Bar{g}_\phi\| \|\Bar{g}_\theta\|} \geq T_\alpha \right\},
\end{equation}
where $T_\alpha$ is determined by a percentile $\alpha$ of trainable parameters with the highest similarity scores between $\Bar{g}_\theta$ and $\Bar{g}_\phi$. Consequently, the optimization in Eq.~\eqref{eq:opt-gd} is reformulated as
\begin{equation}\label{eq:opt-mask}
    \pi_\theta \leftarrow \pi_\theta - M_{\Tilde{g}_\theta} \cdot \left(\nabla_\theta\mathcal{L}_{vl}(\theta) + \nabla_\theta\mathcal{L}_v(\theta)\right).
\end{equation}
We summarize and illustrate the optimization process of MDGD and MDGD-GM in Algorithm~\ref{alg}.

\begin{table*}[ht]
  \centering
  \small
   \resizebox{1.\textwidth}{!}{%
  \begin{tabular}{lcccccccccc}
    \toprule
    \multirow{2}{*}{Method} & \multirow{2}{*}{\#Params} & \multicolumn{6}{c}{Pre-trained tasks} & \multicolumn{1}{c}{Target task} & \multicolumn{2}{c}{Metrics} \\
    \cmidrule(lr){3-8} \cmidrule(lr){9-9}  \cmidrule(lr){10-11}
                    &        & \textbf{GQA}              & \textbf{VizWiz}                & \textbf{SQA}                   & \textbf{TextVQA} & \textbf{POPE} & \textbf{MMBench} & \textbf{Flickr30k} & \textbf{Avg} & \textbf{Hscore} \\
    \midrule
    \textbf{Zero-shot}       & --      & 61.94            & 50.00                 & 66.80                 & 58.27 & 85.90 & 64.30 & 3.5 & 55.82  & 59.86 \\
    \midrule
    \textbf{Fine-tune}       & 1.2B   & 56.26            & 44.45                 & 28.34                 & 38.98 & 38.40 & 50.56 & \textbf{78.82} & 47.97 & 45.26 \\
    \textbf{LoRA}            & 29M   & 17.74            & 40.63                 & 5.38                  & 30.48 & 2.40  & 9.55  & 64.18 & 24.33 & 20.49 \\
    \textbf{Model Tailor}    & 273M   & 52.49            & 42.28                 & \underline{67.15}     & 43.89 & 82.88 & 63.40 & \underline{75.40} & 61.07 & 59.85 \\
    \midrule
    \textbf{MDGD}             & 1.2B   & \underline{67.71}  & \underline{48.18} & \textbf{69.05}         & \underline{57.32} & \textbf{85.12} & \underline{65.43} & 73.47 & \textbf{66.61} & \textbf{66.03} \\
    ~~w/o visual align     & 1.2B   & 57.64           & 36.95                 & 53.96                 & 32.84 & 30.43 & 56.66 & 65.58 & 47.72 & 46.19 \\
    \textbf{MDGD-GM } & 124M   & \textbf{69.89}  & \textbf{51.22}        & 65.87                 & \textbf{58.18} & \underline{84.39} & \textbf{66.42} & 64.18  & \underline{65.74} & \underline{65.86} \\
    \bottomrule
    \toprule

    \multirow{2}{*}{Method} & \multirow{2}{*}{\#Params} & \multicolumn{6}{c}{Pre-trained tasks} & \multicolumn{1}{c}{Target task} & \multicolumn{2}{c}{Metrics} \\
    \cmidrule(lr){3-8} \cmidrule(lr){9-9}  \cmidrule(lr){10-11}
    &                       & \textbf{GQA} & \textbf{VizWiz} & \textbf{SQA} & \textbf{TextVQA} & \textbf{POPE} & \textbf{MMBench} & \textbf{OKVQA} & \textbf{Avg} & \textbf{Hscore} \\
    \midrule
    \textbf{Zero-shot}       & --     & 61.94 & 50.00 & 66.80 & 58.27 & 85.90 & 64.30 & 0.14 & 55.34 & 59.58 \\
    \midrule
    \textbf{Fine-tune}     & 1.2B  & 62.98 & 40.59 & 59.84 & 48.38 & 71.42 & 51.98 & \underline{69.10} & 57.76 & 56.79 \\
    \textbf{LoRA}            & 29M   & 63.44 & 41.61 & 51.29 & 48.02 & 75.27 & 37.31 & \textbf{71.46} & 55.49 & 54.12 \\
    \textbf{Model Tailor}    & 273M   & 60.39              & \textbf{46.49} & \textbf{69.51} & \textbf{54.88} & \textbf{85.44} & \underline{63.32} & 38.10 & 59.73 & 61.48 \\
    \midrule
    \textbf{MDGD}          & 1.2B     & \textbf{66.55}     & 42.72 & 64.60 & 52.54 & \underline{85.17} & 61.73 & 62.29 & \underline{62.23} & \underline{62.22} \\
    ~~w/o visual align & 1.2B  & \underline{66.39} & 39.89 & 60.19 & 52.40 & 84.92 & 62.97 & 62.39 & 61.31 & 61.22 \\
    \textbf{MDGD-GM}  & 124M  & 66.02              & \underline{43.97} & \underline{67.91} & \underline{52.80} & 84.70 & \textbf{63.97} & 61.04 & \textbf{62.92} & \textbf{63.07} \\
    \bottomrule
  \end{tabular}
  }
  \caption{
  Performance on various pre-trained tasks of LLaVA-1.5 models fine-tuned on Flickr30K and OKVQA. 
  We report the best performance for each task in a \textbf{bold font} while the second best performance \underline{underlined}. 
  }
  \label{tab:llava-main}
\end{table*}
\section{Experiments}
\noindent\textbf{Datasets}
To evaluate catastrophic forgetting, we follow the setup of \citet{zhu2024model} and consider two models: LLaVA-1.5 (7B) and MiniCPM-V-2.0 (2.8B). For each model, we use a set of pre-trained tasks (including VQAv2, GQA, VizWiz, TextVQA, POPE, and MM-Bench) and fine-tuning tasks on previously unseen datasets (such as Flickr30k, OKVQA, TextCaps, and PathVQA). Full details on dataset composition are provided in the appendix.

\noindent\textbf{Baselines}
We compare our approach against several baselines: standard fine-tuning following \citet{zhu2024model}, LoRA-based fine-tuning \citep{hu2021lora}, and Model Tailor \citep{zhu2024model}. For Model Tailor, we report the original results from their paper for comparison.

\noindent\textbf{Implementation Details}
All experiments use the official Huggingface implementations of LLaVA-1.5 and MiniCPM-V-2.0, with LoRA adapters where applicable. Models are fine-tuned using BFloat16 precision on 2 NVIDIA A100 80GB GPUs.
We include fine-grained implementation details in Appendix~\ref{app:implement}.

\noindent\textbf{LLM Usage}
In this paper, LLMs are only used for refining the writing of natural language. 

\subsection{Comparison Results}\label{sec:main-results}
\noindent\textbf{LLaVA-1.5 adapts better to downstream tasks but is more prone to visual forgetting.}
We study the visual forgetting problem on the LLaVA-1.5 MLLM,
and report performance comparison results in Table~\ref{tab:llava-main}.
We observe that the pre-trained LLaVA enables efficient instruction tuning on target tasks,
where the zero-shot performance is near zero.
When the model is fine-tuned on the image caption task, Flickr30K, which largely differs from the pre-trained tasks of visual question-answering,
the model can learn a degraded multimodal representation,
which causes visual forgetting in its projected visual representation space (in Section~\ref{sec:visual_forget}).
Fine-tuning on visual question-answering task OKVQA, which is similar to the pre-trained tasks, 
can also lead to MLLM's visual understanding drift, due to the limited image-text pairs existing in the downstream task.

\begin{table*}[ht]
  \centering
  \small
\resizebox{1.\textwidth}{!}{%
  \begin{tabular}{lccccccccccc}
    \toprule   
    \multirow{2}{*}{Method} & \multirow{2}{*}{\#Params} & \multicolumn{7}{c}{Pre-trained tasks} & \multicolumn{1}{c}{Target task} & \multicolumn{2}{c}{Metrics} \\
    \cmidrule(lr){3-9} \cmidrule(lr){10-10} \cmidrule(lr){11-12}
    &                       & \textbf{VizWiz} & \textbf{A-OKVQA} & \textbf{OKVQA} & \textbf{TextVQA} & \textbf{IconQA} & \textbf{POPE} & \textbf{MMBench} & \textbf{PathVQA} & \textbf{Avg} & \textbf{Hscore} \\
    \midrule
    \textbf{Zero-shot}       & -     & 55.27  & 79.39 & 64.86 & 77.98 & 79.01 & 88.93 & 70.98 & 5.44 & 65.23   & 10.04\\
    \midrule
    \textbf{Fine-tune}       & 517M  & 52.91  & 76.94 & 59.06 & 58.34 & 76.96 & \textbf{89.60} & 70.16 & \underline{11.04} & 61.88  & \underline{18.74}\\
    \textbf{LoRA}            & 35M & 52.95  & 76.24 & \textbf{64.45} & 77.18 & 77.80 & 88.08 & 67.47 & \textbf{15.03} & 64.90 & \textbf{24.41}\\
    \midrule
    \textbf{MDGD}           &  517M  & \textbf{55.73}  & 78.25 & \underline{64.33} & \underline{77.54} & \textbf{79.45} & \underline{89.19} & \textbf{71.94} & 9.09 & \textbf{65.69}   & 15.97\\
    ~~w/o visual align & 517M  & 54.92  & \underline{78.52} & 64.17 & 77.42 & \underline{79.37} & 89.10 & 70.96 & 8.49 & \underline{65.37}  & 15.03\\
   \textbf{ MDGD-GM}  & 52M  & \underline{55.04}  & \textbf{78.78} & 64.31 & \textbf{77.78} & 79.10 & 88.76 & \underline{70.98} & 5.72 & 65.06  & 10.52 \\
    
    \bottomrule
    \toprule
    
    \multirow{2}{*}{Method} & \multirow{2}{*}{\#Params} & \multicolumn{7}{c}{Pre-trained tasks} & \multicolumn{1}{c}{Target task} & \multicolumn{2}{c}{Metrics} \\
    \cmidrule(lr){3-9} \cmidrule(lr){10-10} \cmidrule(lr){11-12}
    &                        & \textbf{VizWiz} & \textbf{A-OKVQA} & \textbf{OKVQA} & \textbf{TextVQA} & \textbf{IconQA} & \textbf{POPE} & \textbf{MMBench} & \textbf{TextCaps} & \textbf{Avg} & \textbf{Hscore} \\
    \midrule
    \textbf{Zero-shot}        & -     & 55.27  & 79.39 & 64.86 & 77.98 & 79.01 & 88.93 & 70.98 & 15.77 & 66.52 & 25.50  \\
    \midrule
    \textbf{Fine-tune}        & 517M  & 52.03  & 77.73 & 59.16 & 67.24 & 78.67 & 88.20 & 71.42 & \textbf{33.85} & 66.04 & \textbf{44.76} \\
    \textbf{LoRA}             & 35M  & 53.30  & \underline{78.17}          & \underline{63.99} & \underline{77.68} & 78.28 & 87.31 & 69.23 & \underline{32.41} & 67.55 & \underline{43.80} \\
    \midrule
    \textbf{MDGD}             & 517M  & \textbf{55.17}  & \underline{78.17} & 63.67          & 76.08 & \underline{79.40} & \textbf{89.11} & \underline{71.58} & 28.90 & \underline{67.76} & 40.52 \\
    ~~w/o visual align & 517M  & 51.35           & 78.08            & 63.06          & 76.48 & 78.99                & \underline{88.98} & 71.30 & 25.93 & 66.77 & 37.35 \\
    \textbf{MDGD-GM}  & 52M   & \underline{55.04}  & \textbf{78.43} & \textbf{65.26} & \textbf{78.08} & \textbf{79.65} & 88.93           & \textbf{71.88} & 29.14 & \textbf{68.30} & 40.85 \\
    \bottomrule
  \end{tabular}
  }
  \caption{
  Performance on various pre-trained tasks of MiniCPM-V2.5 models fine-tuned on PathVQA and TextCaps. 
  We report the best performance for each task in a \textbf{bold font} while the second best performance \underline{underlined}.
  }
  \label{tab:minicpm}
\end{table*}

\noindent\textbf{MiniCPM-V-2.0 also experiences visual forgetting while limited in downstream task improvements.}
To validate the observation on a smaller MLLM, we report the comparison results of MiniCPM-V-2.0 with 2.8B model parameters in Table~\ref{tab:minicpm}.
We observe that compared with the LLaVA MLLM, MiniCPM suffers from less prominent visual forgetting.
We attribute this observation to MiniCPM learning a more compact and constrained visual representation space during pre-training, 
causing the visual representations of target task images to be less aligned with those of the pre-trained MLLM. 
Consequently, MiniCPM exhibits limited improvement in downstream tasks, 
as its restricted ability to acquire additional visual knowledge leads to ineffective instruction tuning.

\noindent\textbf{MDGD prevents visual forgetting while maintaining downstream task improvements.}
By employing MDGD in MLLM instruction tuning,
we observe the LLaVA's average performance drops on pre-trained tasks when fine-tuned on OKVQA
and also improves when fine-tuned on Flickr30K,
which demonstrates the efficiency of MDGD in mitigating visual forgetting.
For the smaller MLLM, MiniCPM, MDGD achieves comparable fine-tuning improvements with direct fine-tuning,
while completely eliminating visual forgetting in the pre-trained tasks.
MDGD and its variants consistently achieve the best average performance for both MLLMs,
demonstrating its great potential for incremental learning on individual downstream tasks.

\noindent\textbf{Comparison with baseline methods.}
Table~\ref{tab:llava-main} shows that MDGD consistently outperforms both LoRA fine-tuning and Model Tailor~\cite{zhu2024model} on LLaVA-1.5.
LoRA suffers from visual forgetting due to projecting multimodal features into lower-rank spaces, 
especially on Flickr30K and OKVQA. Model Tailor,
while effective for anti-forgetting in LLMs, 
is less robust for MLLMs and remains sensitive to the target dataset,
performing better on Flickr30K than OKVQA. 
In contrast, MDGD achieves higher average scores and H-scores across datasets.
In Table~\ref{tab:minicpm}, MDGD improves average performance on MiniCPM tasks, reducing visual forgetting by 2.43\% and 1.83\% on PathVQA and TextCaps, respectively.

\begin{figure*}[htp]
    \centering
    \subfigure[$z^{vl}$ on \textbf{PathVQA}]{\includegraphics[width=0.24\textwidth]{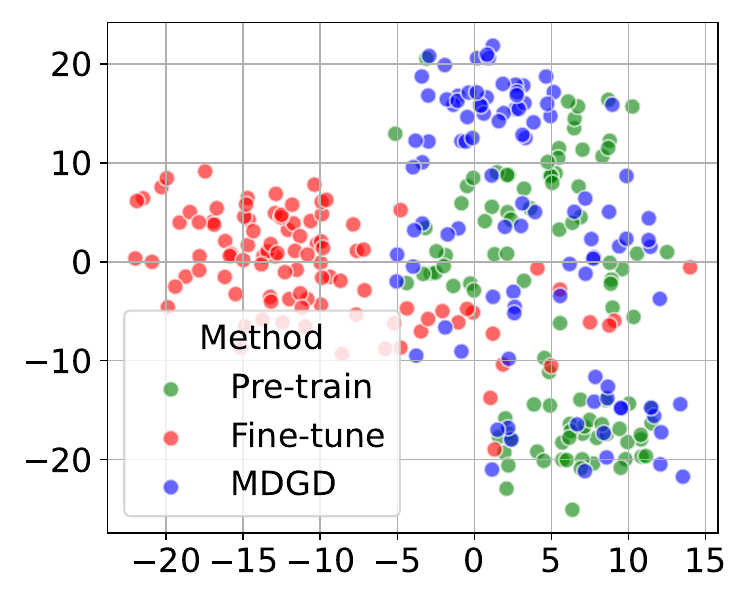}}
    \subfigure[$\pi(X^v)$ on \textbf{PathVQA}]{\includegraphics[width=0.24\textwidth]{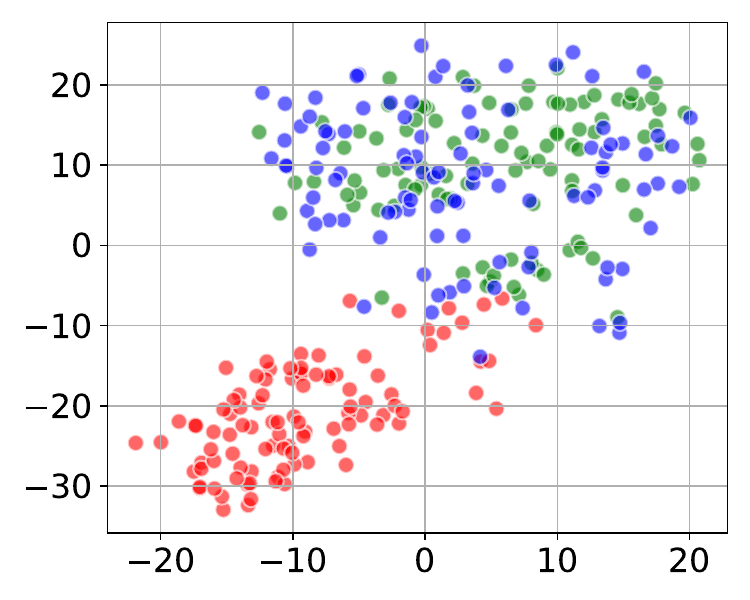}}
    \subfigure[$z^{vl}$ on \textbf{TextCaps}]{\includegraphics[width=0.24\textwidth]{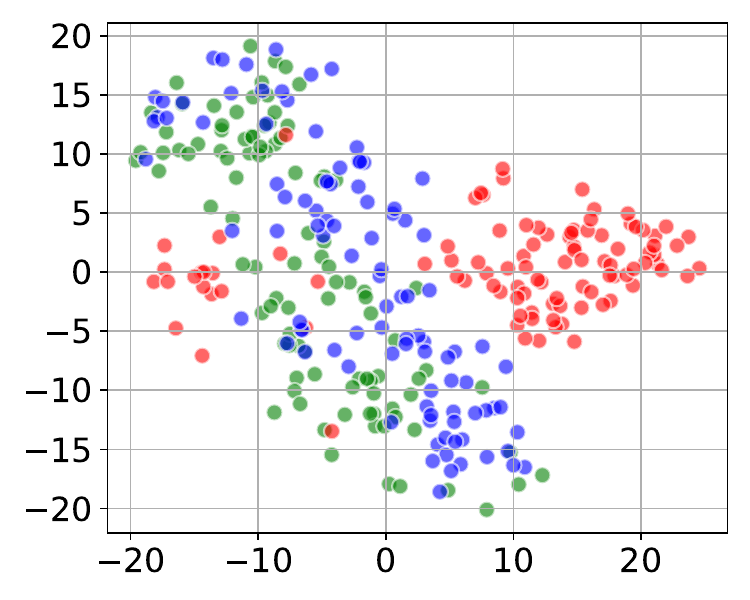}} 
    \subfigure[$\pi(X^v)$ on \textbf{TextCaps}]{\includegraphics[width=0.24\textwidth]{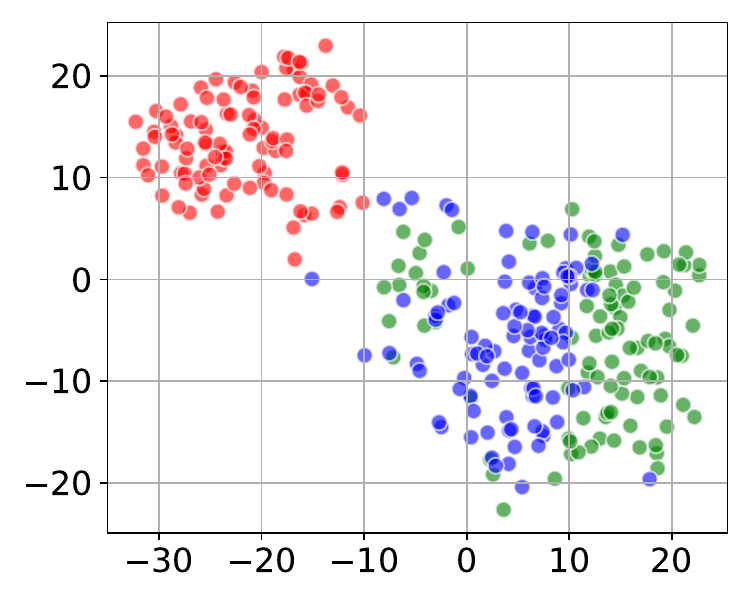}}
    
    \caption{
    T-SNE plots of the distribution of extracted visual $\pi(X^v)$ and multimodal $z^{vl}$ representations from pre-trained MiniCPM, 
    and models with direct fine-tuning and MDGD on PathVQA and TextCaps.
    }
    \label{fig:tsne-cpm}
    \vspace{-1em}
\end{figure*}
\begin{figure*}[htp]
    \centering
    \includegraphics[width=1.\linewidth]{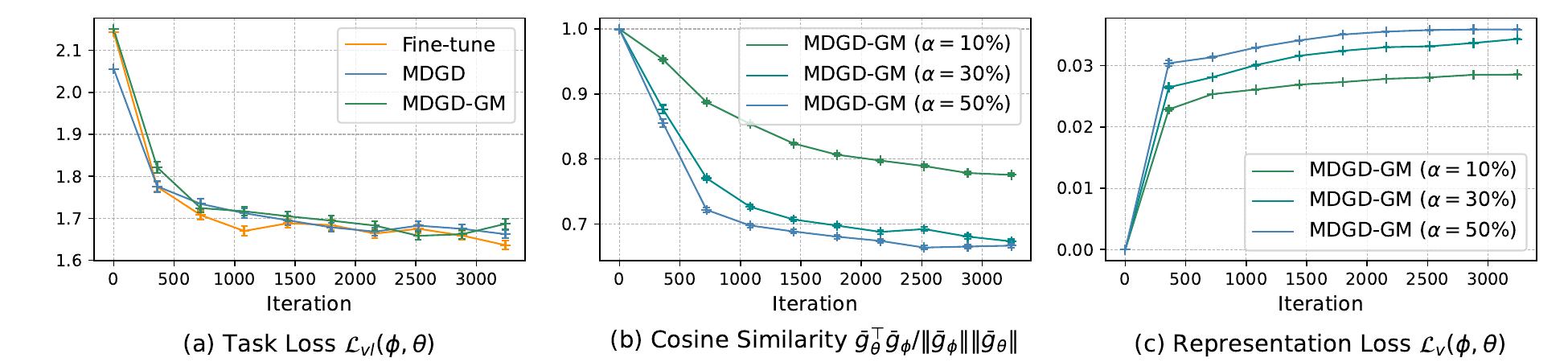}
    \caption{
    Illustration of (a) the learning process of three methods based on task loss $\mathcal{L}_{vl}(\phi,\theta)$, 
    (b) the average regularized cosine similarity $\frac{\Bar{g}_\theta^\top \Bar{g}_\phi}{\|\Bar{g}_\phi\| \|\Bar{g}_\theta\|}$ in Eq.\eqref{eq:masking} for gradient masking at varying ratios, 
    and (c) the visual representation loss $\mathcal{L}_v(\phi,\theta)$ in Eq.\eqref{eq:visual_loss} for gradient masking at varying ratios $\alpha$.
    } 
    \label{fig:learning}
    \vspace{-.6cm}
\end{figure*}

\subsection{Ablation Study}
\noindent\textbf{Ablation study on visual alignment.}
We compare MDGD with its two variants, MDGD w/o visual align and MDGD-GM.
MDGD w/o visual align enables MDGD without including visual representation loss $\mathcal{L}_v(\phi,\theta)$ Eq.\eqref{eq:visual_loss}, 
to understand the effect of directly optimizing to reduce the visual representation discrepancy between the current model and pre-trained model.
We observe that MDGD w/o visual align maintains relatively comparable  performance to MDGD on OKVQA and PathQA,
due to the reduced need for visual representation adaptation in such visual question-answering tasks.
In contrast, tasks like image captioning on Flickr30K and TextCaps benefit from feature alignment regularization, 
which directly mitigates visual understanding drift in the MLLM.

\noindent\textbf{Ablation study on gradient masking.}
The other variant, MDGD-GM, leverages gradient masking to enable parameter-efficient fine-tuning (PEFT).
We observe the PEFT variant of MDGD consistently achieves comparable performance across all tasks and backbone MLLMs,
which only fine-tunes a subset of 10\% original MLLM parameters used for direct fine-tuning and original MDGD. 
Different from conventional PEFT methods such as adapters, 
MDGD and its variants do not introduce additional parameters to the original model architecture, 
enabling incremental learning in an online setting \citep{maltoni2019continuous,gao2023llama}.

\subsection{Representation Learning Analysis} \label{sec:repre}
\noindent\textbf{T-SNE Analysis on Visual Representation}
To analyze the learning of visual and multimodal representation distributions in MLLMs, 
we create T-SNE \cite{van2008visualizing} plots to visualize the feature distributions extracted from pre-trained MLLMs, 
as well as MLLMs after standard fine-tuning and MDGD
We illustrate the distributions of the multimodal features $z^{vl}$ extracted from the last token of the multimodal instruction tokens, 
and the visual features $\pi_\theta(X^v)$ extracted from the last token of the input image tokens.
We observe a consistent visual understanding drift in the MLLMs' visual representation spaces after standard fine-tuning on PathVQA and TextCaps with MiniCPM (Figure~\ref{fig:tsne-cpm}b and \ref{fig:tsne-cpm}d).
By employing MDGD to mitigate visual forgetting, we observe that visual understanding drift is effectively reduced, 
allowing the fine-tuned MLLM to retain pre-trained visual capabilities.

We further observe a distributional discrepancy in the multimodal 
representation $z^{vl}$ of LLaVA (Figures~\ref{fig:tsne-llava}a and \ref{fig:tsne-llava}c) 
between MDGD and the pre-trained MLLM. 
This discrepancy arises from the alignment of the MLLM to the target task through multimodal instructions, 
demonstrating effective adaptation to the downstream task of the LLaVA model.
In addition, we also observe such multimodal distribution discrepancy reduces in a smaller MLLM, MiniCPM.
This observation aligns with our findings on MiniCPM in Section~\ref{sec:main-results}, 
where we noted limited effects in model adaptation to downstream tasks. 
However, applying MDGD to MiniCPM mitigates visual forgetting by preventing degradation of both image and multimodal encodings into lower-rank representation spaces.

\begin{figure}[htp]
    \centering
    \subfigure[LLaVA models pretrained, finetuned, and fine-tuned with MDGD]{%
        \includegraphics[width=1.\linewidth]{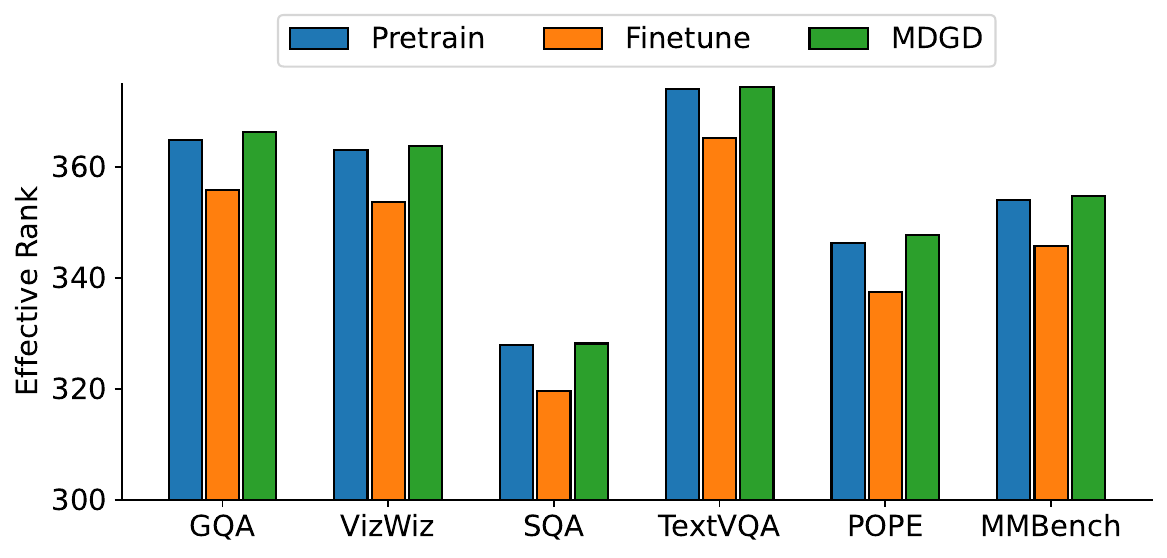} 
        \label{fig:erank-eval1}
    }
    \hfill
    \subfigure[MiniCPM models pretrained, finetuned, and fine-tuned with MDGD]{%
        \includegraphics[width=1.\linewidth]{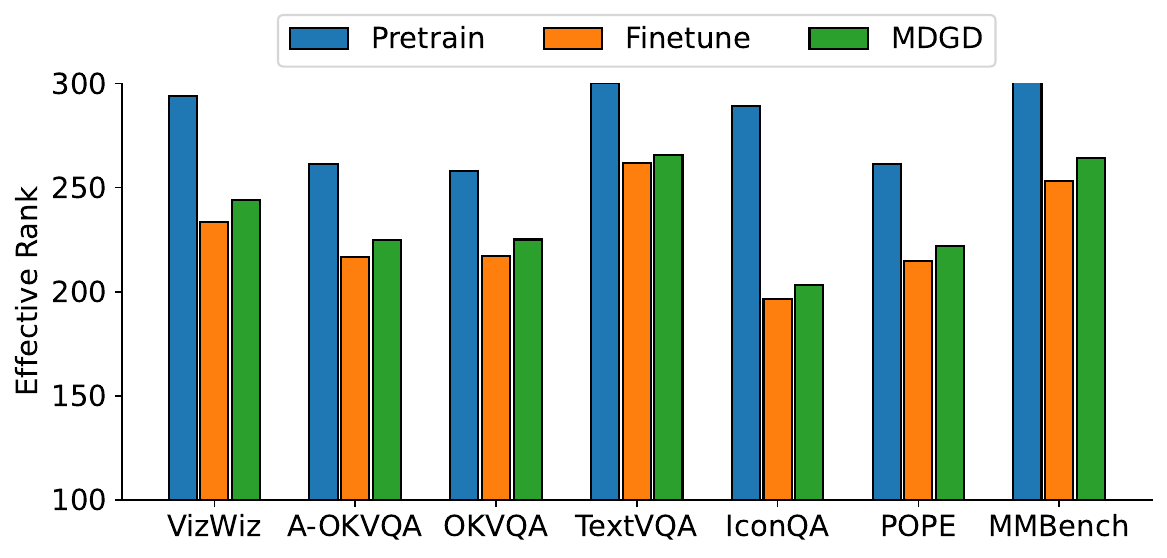} 
        \label{fig:erank-eval2}
    }
    
    \caption{The effective rank comparison on individual downstream fine-tuning datasets.}
    \label{fig:erank-eval}
    \vspace{-1em}
\end{figure}

\noindent\textbf{Effective Rank Analysis on Visual Representation}
To quantitatively analyze the visual forgetting problem (in Section~\ref{sec:visual_forget}) described in Eq.~\eqref{eq:erank_degradation},
we calculate effective ranks of the visual representations extracted from the last hidden layer on the position of image tokens in individual MLLMs.
We show the comparison results of LLaVA models in Figure~\ref{fig:erank-eval1} and MiniCPM models in Figure~\ref{fig:erank-eval2}.
We observe that with both the backbone models of LLaVA and MiniCPM, 
directly fine-tuning the pre-trained models on downstream tasks can lead to a consistent reduction of effective ranks in visual representations.
Such observations validate the hypothesis in Section~\ref{sec:visual_forget} regarding the potential visual forgetting problem in MLLM instruction tuning.
In addition, we can observe that MDGD achieves consistent improvements in effective ranks compared with the standard fine-tuning method for both backbone MLLMs across various pre-trained tasks.
In Figure~\ref{fig:erank-eval1}, we observe that MDGD achieves comparable or even better effective ranks on pre-trained tasks, compared with the pre-trained LLaVA model.
However, MDGD on MiniCPM in Figure~\ref{fig:erank-eval2} also suffers from the visual representation degradation problem, while MDGD consistently alleviates the problem.
Such observation suggests a higher risk of visual forgetting in smaller-scale MLLMs.

\subsection{Sensitivity Study}
We evaluate the learning curves of MDGD and MDGD-GM compared with standard fine-tuning in Figure~\ref{fig:learning}(a),
where we observe that MDGD and MDGD-GM achieve comparable training efficiency compared with the standard fine-tuning method.
We also investigate the sensitivity of gradient cosine similarity between $\Bar{g}_\theta$ and $\Bar{g}_\phi$ in Figure~\ref{fig:learning}(b) and the representation loss in Figure~\ref{fig:learning}(c),
with respect to the gradient masking ratio in MDGD-GM.
In Figure~\ref{fig:learning}(b), we observe that MDGD-GM with lower gradient masking ratios can better align the modality-decoupled learning gradients between the target model and the pre-trained model,
while MDGD-GM maintains over 70\% alignment with 50\% gradient masking.
In Figure~\ref{fig:learning}(c), we show that MDGD-GM with 50\% gradient masking still effectively alleviates the visual representation degradation problem by reducing the visual representation discrepancy $\mathcal{L}_v$,
while learning with a more active gradient can achieve better alignment.
\section{Conclusion}
In this work, we addressed the challenge of visual forgetting in MLLMs during instruction tuning by introducing a novel modality-decoupled gradient descent (MDGD) approach. MDGD disentangles the gradient updates for visual representation learning from task-specific alignment, thereby preserving the effective rank of pre-trained visual features and mitigating the over-compression effects highlighted by the information bottleneck perspective. This decoupling enables MLLMs to retain rich visual knowledge while adapting robustly to new downstream tasks. Furthermore, our gradient masking variant, MDGD-GM, enhances memory efficiency and optimizes parameter usage, making fine-tuning both practical and scalable. Extensive experiments across various downstream tasks and backbone models demonstrate that MDGD not only effectively prevents visual forgetting but also outperforms existing strategies in achieving balanced multimodal representation learning and task adaptation. Our findings underscore the importance of preserving visual representations during instruction-tuning and offer a viable solution for efficient and effective multimodal learning in real-world scenarios.

\section{Limitation}
In this work, we focus on MLLMs that process multimodal instructions consisting solely of visual and textual inputs. Given the limited availability of MLLMs across other modalities, our primary goal is to mitigate visual forgetting. However, our modality-decoupling approach is generalizable to other input modalities. Consistent with standard practices, we limit the instructions to two input modalities, though extending this to more diverse, free-form multimodal inputs remains an avenue for future research.
\section*{Acknowledgment}
This work is partially supported by NSF IIS-2432486.

\bibliography{main}

\appendix

\section{T-SNE Analysis on LLaVA-1.5 Model}
\begin{figure*}[htp]
    \centering
    \subfigure[$z^{vl}$ on \textbf{OKVQA}]{\includegraphics[width=0.24\textwidth]{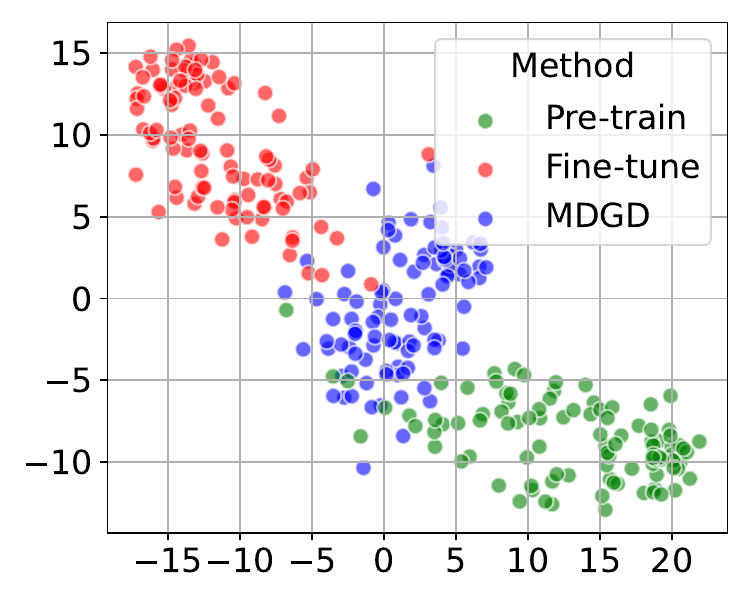}}
    \subfigure[$\pi(X^v)$ on \textbf{OKVQA}]{\includegraphics[width=0.24\textwidth]{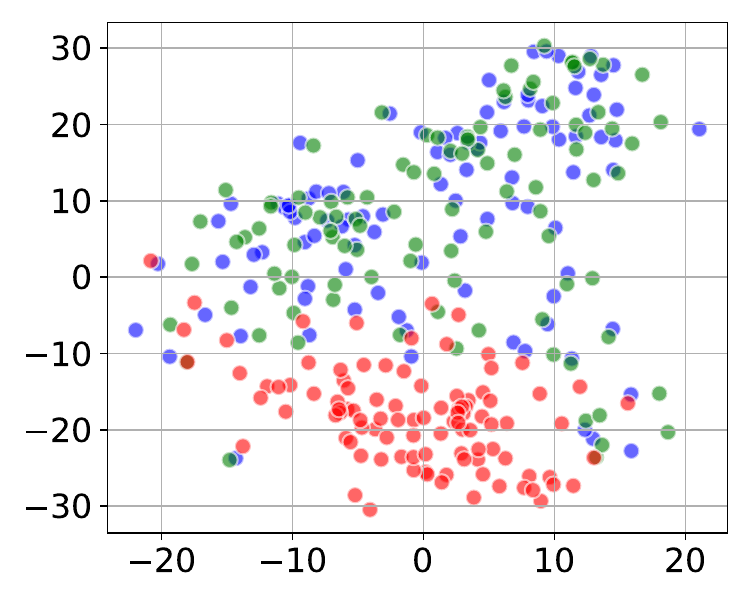}}
    \subfigure[$z^{vl}$ on \textbf{Fllickr30K}]{\includegraphics[width=0.24\textwidth]{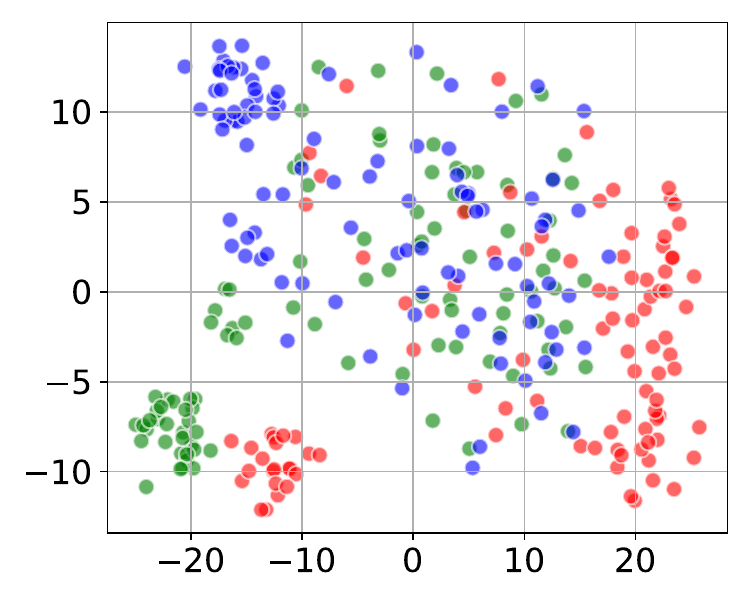}} 
    \subfigure[$\pi(X^v)$ on \textbf{Fllickr30K}]{\includegraphics[width=0.24\textwidth]{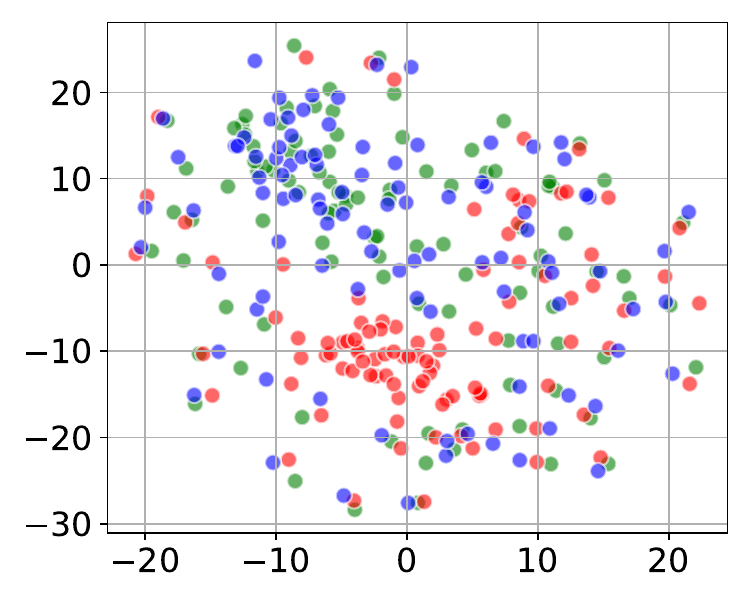}}
    
    \caption{
    T-SNE plots of the distribution of extracted visual $\pi(X^v)$ and multimodal $z^{vl}$ representations from pre-trained LLaVA-1.5, 
    and models with direct fine-tuning and MDGD on OKVQA and Flickr30K.
    }
    \label{fig:tsne-llava}
  \vspace{-1em}
\end{figure*}

In addition to Section~\ref{sec:repre}, we further include the T-SNE analysis on LLaVA-1.5 model. 
We observe a consistent visual understanding drift in the MLLMs' visual representation spaces after standard fine-tuning on Flickr30K and OKVQA with LLaVA (Figure~\ref{fig:tsne-llava}b and \ref{fig:tsne-llava}d).

\section{Implementation Details} \label{app:implement}

\noindent\textbf{Datasets}
To evaluate the effectiveness of MDGD in mitigating catastrophic forgetting, we used two models of different sizes. 
Our experimental design follows the settings from the work of \citet{zhu2024model}. 
For each model, datasets were categorized into two types: 
\textbf{pre-trained tasks}, which assess the model's ability to retain inherent knowledge after fine-tuning, 
and \textbf{fine-tuning tasks}, consisting of unseen datasets used to test adaptability. 
After fine-tuning, we evaluated performance on both task types to measure forgetting and generalization. 
Below, we detail the datasets used for each model. 
\textbf{LLaVA-1.5 (Vicuna-7B) \citep{liu2024improved}}: This model has 7 billion parameters. In line with \citet{liu2024improved}, we used the following datasets:
\begin{itemize}
    \item \textbf{Pre-trained Tasks}: VQAv2 \citep{goyal2017making}, GQA \citep{hudson2019gqa}, VizWiz \citep{gurari2018vizwiz}, SQA \citep{lu2022learn}, TextVQA \citep{singh2019towards}, POPE \citep{li2023evaluating}, and MM-Bench \citep{liu2023mmbench}.
    \item \textbf{Fine-tuning}: Flickr30k \citep{young2014image} and OKVQA \citep{marino2019ok}, which were not encountered in the pre-training stage.
\end{itemize}
\textbf{MiniCPM-V-2.0 \citep{yao2024minicpmvgpt4vlevelmllm}}: This model has 2.8 billion parameters. We evaluated its performance on:
\begin{itemize}
    \item \textbf{Pre-trained Tasks}: VizWiz, OKVQA, A-OKVQA \cite{schwenk2022okvqa}, Text-VQA, IconQA \cite{lu2021iconqa}, POPE, and MM-Bench.
    \item \textbf{Fine-tuning}: TextCaps \citep{sidorov2020textcaps} and PathVQA \citep{he2020pathvqa}, which were not part of its pre-training exposure.
\end{itemize}

\noindent\textbf{Baselines} We compare our approach against several baselines: 
\begin{itemize}
\item \textbf{Standard Fine-Tuning.} For a fair comparison, we follow the setting of Model-Tailor \cite{zhu2024model}, where LLaVA-1.5 is fine-tuned on the last 6 layers and its feature adapter, with a total of 1.2B parameters.
MiniCPM is fine-tuned on the last 8 layers and its feature resampler, with 517M parameters. 
\item \textbf{LoRA-based Fine-Tuning \citep{hu2021lora}.} LoRA introduces low-rank matrices to update only a small subset of parameters, reducing memory consumption and computational cost. In our experiments, LLaVA-1.5 and MiniCPM are fine-tuned by modifying the query and key projection layers within the attention mechanism. 
\item \textbf{Model Tailor \citep{zhu2024model}.} This baseline employs a hybrid strategy that mitigates catastrophic forgetting by identifying and adjusting the most critical parameters for adaptation. It has been evaluated through experiments on multimodal large language models (MLLMs). As the method is not open source, we report only the original results of the LLaVA-1.5 experiments provided in the original paper as a baseline.
\end{itemize}

\noindent\textbf{Implementation Details}
We use the official Huggingface implementations of the LLaVA-1.5 and the MiniCPM-V-2.0 models and their LoRA adapters. 
For model fine-tuning, we use BFloat16 precision for memory-efficient training. 
Experiments are conducted using 2 NVIDIA A100-SXM4-80GB GPUs.

\section{Visual Forgetting in MLLM Instruction-tuning}\label{sec:visual_forget}
Building on the IB objective Eq. (\ref{eq:ib_vision}) introduced in Section~\ref{sec:prelim}, we examine how instruction tuning affects the richness of visual representations. 
Let the pre-trained MLLM induce a latent representation,  
$$Z \sim p(\cdot\mid X^v, X^l),$$
where $Z$ is is decomposed into modality-specific components, $Z = (Z^v, Z^l)$
with \( Z^v \) captures the visual features extracted from \( X^v \), and \( Z^l \) encapsulates the textual features from \( X^l \).
Define the \emph{pre-trained} visual representation space as,
\[
\mathcal{Z}_0^v = \left\{ Z^v_\phi : Z \sim p_\phi(\cdot \mid X^v), \quad X^v \in \Omega \right\}.
\]
During instruction tuning, the model is optimized primarily to predict the target $y$. As described in Eq. (\ref{eq:ib_vision}), the IB objective introduces a trade-off between retaining visual information \( I(X^v; Z) \) and ensuring that \( Z \) remains predictive of \( y \) via \( I(y; Z) \)~\cite{jiang2024correlation}. 
In practice, however, instruction-tuning datasets are predominantly text-driven; thus, the learned visual representation $Z^v$ receives only indirect and often weaker supervision~\cite{wang2024mdpo}.

Let the tuned model’s latent representation be $Z_\theta \sim p_\theta(\cdot\mid X^v, X^l),$
and denote the corresponding visual representation space by,
\[
\mathcal{Z}_\theta^v = \left\{ Z^v_\theta : Z \sim p_\theta(\cdot\mid X^v, X^l), (X^v, X^l) \in D \right\},
\]
where \( D \) is the instruction-tuning dataset.
To measure the richness of the visual representation, we employ the effective rank metric from Eq.~(\ref{eq:erank}). 
A higher effective rank indicates that the representation spans a broader subspace, whereas a lower effective rank signals more aggressive compression.  

\vspace{1em}\noindent\textbf{The Visual Forgetting Problem.} 
During instruction tuning, the visual representation undergoes significant compression as the model prioritizes textual supervision. 
This reduction occurs because the model effectively sacrifices part of $I(X^v; Z)$ to focus on $I(y; Z)$, thereby reducing the effective dimensionality of the visual features. 
As a result, the model progressively loses its ability to retain and utilize rich visual information, leading to a phenomenon we define as \textbf{\textit{visual forgetting}}.
Empirically, in Figure~\ref{fig:intro} we observe,
\begin{equation}\label{eq:erank_degradation}
    \text{erank}(\mathcal{Z}_\theta^v) < \text{erank}(\mathcal{Z}_0^v).
\end{equation}
This indicates that the tuned visual representation is compressed relative to the pre-trained space, making it harder for the model to leverage visual information effectively.
In RQ3 (Section~\ref{sec:repre}), we validate such empirical observations and demonstrate that our method helps to preserve effective ranks in the visual representation learning of MLLMs.

\end{document}